\documentclass[10pt,twocolumn,letterpaper]{article}

\usepackage{cvpr}
\usepackage{times}
\usepackage{epsfig}
\usepackage{graphicx}
\usepackage{amsmath}
\usepackage{amssymb}

\usepackage{booktabs} % nice tables
\usepackage{paralist}
\usepackage{comment}
\usepackage{color,colortbl}
\usepackage{tabularx}

\usepackage[utf8x]{inputenc} 

\newcommand{\specialcell}[2][b]{%
  \begin{tabular}[#1]{@{}c@{}}#2\end{tabular}}
\newcolumntype{L}{>{\raggedright\arraybackslash}X}

\usepackage[pagebackref=true,breaklinks=true,letterpaper=true,colorlinks,bookmarks=false]{hyperref}

\cvprfinalcopy % *** Uncomment this line for the final submission

\def\cvprPaperID{***} % *** Enter the CVPR Paper ID here

\ifcvprfinal\fi
\begin{document}

%%%%%%%%% TITLE
\title{\LaTeX\ Author Guidelines for CVPR Proceedings}

\title{DAL -- A Deep Depth-aware Long-term Tracker}

\author{Yanlin Qian$^{1}$, Alan Lukežič$^{2}$, Matej Kristan$^{2}$,Joni-Kristian K\"am\"ar\"ainen$^{1}$,Jiří Matas$^{3}$ \\
{\small $^1$Computing Sciences, Tampere University~~}
{\small $^2$~Faculty of Computer and Information Science, University of Ljubljana, Slovenia}  \\
{\small $^3$~Faculty of Electrical Engineering, Czech Technical University in Prague, Czech Republic}  \\
{\tt\small yanlin.qian@tuni.fi}
}

\maketitle
%\thispagestyle{empty}

%%%%%%%%% ABSTRACT
\begin{abstract}
The best RGBD trackers provide high accuracy but are slow
to run. On the other hand, the best RGB trackers are fast
but clearly inferior on the RGBD datasets.
In this work, we propose a deep depth-aware long-term tracker that achieves state-of-the-art RGBD tracking performance and is fast to run.

We reformulate deep discriminative correlation filter (DCF) to embed the depth information into deep features. Moreover, the same depth-aware correlation
filter is used for target re-detection. 
Comprehensive evaluations show that the proposed tracker achieves state-of-the-art performance on the Princeton RGBD, STC, and the newly-released CDTB benchmarks and runs 20 fps.

\end{abstract}

%%%%%%%%% BODY TEXT
%%%%%%%%%%%%%%%%%%%%%%%%%%%%%%%%%%%%%%%%%%%%%%%%%%%%%%%%%%
\section{Introduction}

Visual object tracking has progressed significantly largely thanks to the series of increasingly challenging visual object tracking benchmarks~\cite{Kristan_VOT2017, Kristan_VOT2018, Kristan_2019_ICCV,GOT10k,TrackingNet, LaSOT}. 
In the most general formulation, a tracker is initialized in the first frame and is required to output the target position in all remaining frames. In many practical applications, such as surveillance systems, trackers need to cope with occlusions
and the target leaving the camera view which are essential
properties for {\em long-term trackers}~\cite{Kristan_VOT2018}.

\begin{figure}[th]
  \begin{center}
    \includegraphics[width=\linewidth]{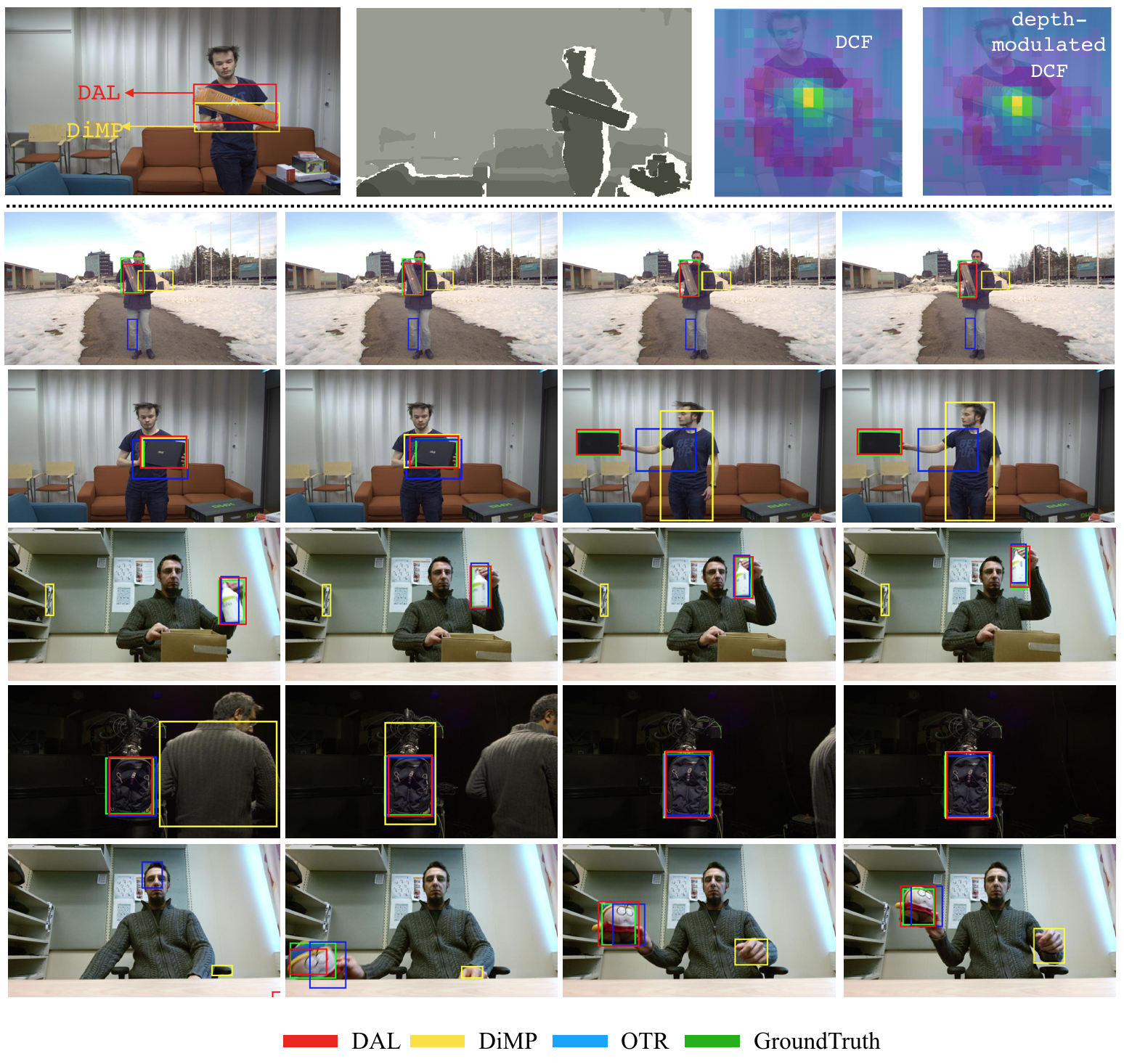}
    \caption{
    Qualitative comparison of our tracker and SoTA RGB and RGBD trackers. 
    The top row illustrates the activation maps from a base DCF and a depth-modulated DCF (better zoom-in to see), generating slightly different shifts of target center and resulting in different bounding boxes (red and yellow). 
    The two videos in the 1st and 2nd rows are non-occlusion scenarios, where our tracker, based on non-stationary DCF, localize the target well while the original DiMP fails, after multiple times of applying DCF operations. 
    In the bottom three rows, the target appears from occlusion and are re-detected and tracked by our long-term tracker in a fast and accurate way.
    }
      \label{fig:introfigure}
  \end{center} 
\end{figure}

A vast majority of the works have focused on RGB tracking,
but recently RGBD (RGB+Depth) tracking has gained momentum. Depth is a particularly strong cue for object's 3D localization, potentially simplifies foreground-background separation for
occlusion handling and even helps to construct a 3D model of the tracked object~\cite{kart2019object}.
Moreover, a number of RGBD datasets
have been introduced in increasing pace~\cite{princetonrgbd,stc,cdtb}.

Recent works~\cite{Kart_ECCVW, kart2019object} have demonstrated improved tracking performance by adopting depth based occlusion handling. However, a recent long-term RGBD tracking benchmark~\cite{cdtb} revealed that the best performance is achieved with the state-of-the-art RGB trackers that omit the depth input. In the most recent RGBD track
of the VOT challenge~\cite{Kristan_2019_ICCV} the best
RGBD trackers outperformed RGB trackers by a clear margin.
These trackers, however, are complicated architectures
using deep object detectors, segmentation and deep
feature based tracker pipelines. Their complex structure makes
them unacceptably slow ($\sim$2fps) for many real-time
applications and it is difficult to improve their
computation without sacrificing accuracy. Speed-wise the best
RGB trackers outperform RGBD trackers, but the
speed-accuracy trade-off gap between the best
RGB and RGBD trackers remains an open problem.

This paper addresses the aforementioned issues and contributes by closing the performance gap in the terms of accuracy and
speed between the RGB and RGBD trackers. We propose a new RGBD tracker of a streamlined architecture that exploits depth information at all levels of processing and obtains performance comparable to the best slow RGBD trackers~\cite{Kristan_2019_ICCV} with the speed comparable to
real-time RGB trackers. The target appearance is modeled
by adopting the state-of-the-art deep discriminative correlation filter (DCF) architecture~\cite{dimp}. However, the
deep DCF is modulated using the depth information such that a large change in the depth suppresses discriminative features in these regions.
The proposed "depth modulated" DCF model performs well
both in short-time frame-to-frame tracking and target
re-detection and therefore makes complex object detection
unnecessary and provides significant speed-up.
Tracking examples are shown in Figure~\ref{fig:introfigure}.

The proposed long-term RGBD tracker achieves state-of-the-art performance on all three available RGBD tracking benchmarks, PTB~\cite{princetonrgbd},  STC~\cite{stc} and CDTB benchmark~\cite{cdtb}, and runs an order of magnitude
faster than the recent state-of-the-art RGBD tracker~\cite{kart2019object} or the winner of the recent
VOT-RGBD challenge~\cite{Kristan_2019_ICCV}. We also
provide an ablation study that confirms the effectiveness of the depth modulated DCF formulation and other components of the
proposed tracker.

\section{Related Work}

\noindent \textbf{RGB trackers.}
Generic visual tracking with RGB input can be roughly divided into two tracks --discriminative correlation filter-based familly (DCF) and Siamese-based familly. Bolme~\etal~\cite{MOSSE2010} inspired the visual tracking community of how visual tracking is addressed by DCF in a mathematical-sound way. DCF was extended by Henriques~\cite{Henriques_KCF} with fourier-transform-based training, and later augmented with segmentation constraints in CSR-DCF~\cite{lukezic2017discriminative}.  

Siamese networks were invented with end-to-end trainable ability and relatively high tracking accuracy~\cite{SiameseFC}. Li~\etal~\cite{SiamRPN} adopts a region proposal network for better predicted bounding boxes. Zhu~\etal~\cite{DaSiamRPN} suppresses the effect of background distractors by controlling the quality of learned target model. The most advanced siamese-based tracker is SiamRPN++\cite{SiamRPN++}, utilizing ResNet-50 for feature representation. Recently, DCF tends to be merged into an end-to-end deep network. The representative work is ATOM~\cite{atom} that allows large-scale training for bounding box estimation and learning discriminative filter on the fly.\\

\noindent \textbf{RGBD trackers.}
There are much less RGD-D trackers, compared to RGB ones. PTB~\cite{princetonrgbd} opened this research topic by presenting a hybrid RGBD tracker composed of HOG feature, optical flow and 3D point clouds. 
Under partical filter framework, Meshgi~\cite{MESHGI_OAPF} addresses RGBD tracker with occlusion awareness and Bibi~\cite{Bibi3D} further models a target using sparse 3D cuboids. Based on KCF, Hannuna~\etal~\cite{Hannuna2016} uses depth for occlusion detection and An~\etal~\cite{DLST} extends KCF with depth channel. Liu~\etal~\cite{ca3dms} presents a 3D mean-shift-based tracker. Kart~\etal~\cite{Kart_ECCVW} applies graph cut segmentation on color and depth information, generating better foreground mask for training CSRDCF~\cite{lukezic2017discriminative}. They then extend the idea with building an object-based 3D model~\cite{kart2019object}, relying on a SLAM system Co-Fusion~\cite{runz2017co}. At the moment of writing this paper, OTR~\cite{kart2019object} leads the leaderboard of two RGBD benchmarks.\\

\noindent \textbf{Benchmarks.} Till now, we briefly introduced the most representative and well-performing RGBD trackers. The reason of their performance lag compared to RGB trackers is obvious -- none of them has access to semantic target-based prior knowledge, which can be obtained via heavy off-line CNN training.  
Tracking benchmarks are crucial for the development of trackers. 

It is obvious that RGBD benchmarks are much smaller than the RGB counterparts by orders of magnitude, for example, the biggest RGB tracking benchmark, TrackingNet~\cite{TrackingNet} contains up to 14 million samples while the biggest RGBD tracking dataset CDTB~\cite{cdtb} 100 thousand samples. Among RGBD datasets, only PTB~\cite{princetonrgbd} provide a tiny subset (hundreds of images) for training, which is far from enough for training or fine-tuning a deep net. 
The shortage of RGBD training set explains why off-line training has not been adopted for RGBD tracking. 
To narrow the performance gap between RGB and RGBD trackers, it is beneficial to use deep features from deep nets pretrained on massive RGB training set. 

\section{Method}  \label{sec:method}

In Sec.~\ref{subsec:robust-localizer}, we first introduce the base RGB tracker briefly. In same section, we also describe the design of depth-aware convolution layer and show its application on DCF-based tracking. In Sec.~\ref{subsec:fine-localizator} we briefly describe the bounding box regressor -- IoUNet. We overview the long-term RGBD tracking architecture in Sec.~\ref{subsec:longterm} , with emphasis on the interaction conditions of switching between short-term tracking and re-detection mode in Sec.~\ref{subsec:target-presence-classifier}.

\subsection{Robust localization}  \label{subsec:robust-localizer}

Robust localization is the most crucial element of long-term tracking. We thus formulate the target model as a deep discriminative correlation filter (DCF), which is trained by the efficient deep training algorithm proposed recently~\cite{dimp}. Given a set of labelled training samples $\mathbf{S}_{test}$, the filter  $\mathbf{f}$ is optimized by steepest descent on the following loss $L_\text{cls}$: 
\begin{equation} \label{eq:classification-loss}
    L_\text{cls} = \frac{1}{N_\text{iter}} \sum_{i=0}^{N_\text{iter}}
    \sum_{(x,c) \in S_\text{test}} 
    \|\ell\big(\mathbf{x} * \mathbf{f}^{(i)}, \mathbf{z}_c\big)\|^2 \,.
\end{equation}
where $*$ is convolution operation and $\mathbf{z}_c$ refers to the corresponding Gaussian function centered to the target location $c$ of the training sample $\mathbf{x}$ and $N_\text{iter}$ is the number of steepest descent iterations. The loss applies a nonlinear regression error $\ell(s, z)=s-z$ for $z>T$ and $\ell(s, z)=\max(0,s)$ for $z \leq T$, where $T$ is a threshold on the error.
The training samples $\mathbf{x} \in S_{test}$ are extracted from the image patch 5 times larger than the target size using a common backbone~\cite{resnet}, which is fine-tuned for localization task~\cite{dimp}.

The target is localized on a new frame by extracting deep features within a patch 5 times the target size and correlated by the trained filter $\mathbf{f}$. Position of the maximum correlation response is the new target position estimate.

\begin{figure}[ht]
  \begin{center}
    \includegraphics[width=\linewidth]{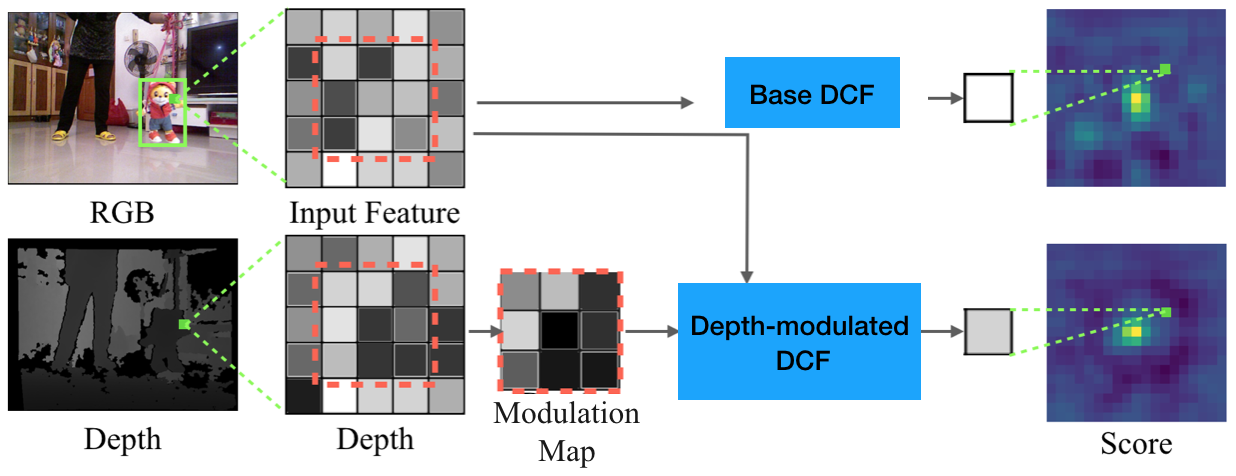}
    \caption{Visualization of depth-modulated DCF. Depth modulates the DCF by re-weighting the DCF kernels according to the depth similarity with the tested target position.  }
      \label{fig:depthawaredcf}
  \end{center}
  \vspace{-5mm}
\end{figure}

However, using a stationary filter (i.e., the same filter) on all locations is sub-optimal since certain regions might contain occlusion and are thus less reliable than other regions~\cite{wang2018depth}. Furthermore, certain targets are poorly approximated by a rectangular convolution window and therefore a mechanism for background suppression is required. To solve all these problems simultaneously, we propose a non-stationary deep DCF that utilizes depth to modulate the DCF content with respect to the filter position. Specifically, we define the new depth-modulated DCF as
\begin{equation}  \label{eq:standard_conv}
    \tilde{\mathbf{f}}(x,y) = \mathbf{f} \odot \mathbf{\Theta}(x,y),
\end{equation}
where $\mathbf{f}$ is a stationary base filter, $\mathbf{\Theta}(x,y)$ is a non-stationary 2D modulation map, and $\odot$ is a Hamadarad product, that multiplies all channels of the base filter with the same modulation map. The purpose of the modulation map is to give more weight to the pixels with depth values similar to the tested target position, thus reducing the effect of the background and  occlusion. Let $\mathbf{D}(x,y)$ be the depth at the tested position and let $\mathbf{D}(x+i,y+j)$ be the depth of the neighboring pixel. The modulation map is then defined as
\begin{equation}  \label{eq:depthdisparity}
    \mathbf{\Theta}_{ij}(x,y) = \exp(-\alpha |\mathbf{D}(x,y) - \mathbf{D}(x+i,y+j)|),
\end{equation}
where $\alpha$ is a hyper parameter that controls the modulation strength.  Figure~\ref{fig:depthawaredcf} illustrates the modulation map construction and usage in non-stationary DCF correlation. The loss for training the non-stationary DCF becomes
\begin{equation} \label{eq:classification-loss-nonsta}
    L_\text{cls} = \frac{1}{N_\text{iter}} \sum_{i=0}^{N_\text{iter}}
    \sum_{(x,c) \in S_\text{test}} 
    \|\ell\big(\mathbf{x} * \tilde{\mathbf{f}}^{(i)}(x,y), z_c\big)\|^2 \, \enspace .
\end{equation}
The loss is optimized using the steepest decent algorithm from~\cite{dimp} within a region five times the target size to harvest a sufficient amount of negative examples. The non-stationary DCF learns to take into account the target-background discontinuities induced by depth and
therefore provides improved foreground-background discrimination.

\subsection{Accurate localization}  \label{subsec:fine-localizator}

The non-stationary depth-modulated DCF described in Section~\ref{subsec:robust-localizer} robustly localizes the target even in presence of clutter.
For accurate bounding box prediction i.e., width and height of the target, we follow the recent IoUNet~\cite{iounet_eccv18} bounding box regression introduced in~\cite{atom}. 

The IoUNet is offline trained on image pairs of the same target using a large number of video sequences.
First image and the corresponding bounding box are used as a training example. 
A modulation vector is extracted from this image and used with the second image (test example) to refine the given test bounding box and to predict its intersection over union with the ground-truth bounding box.

During tracking, after the target is approximately localized by the depth-modulated DCF (Section~\ref{subsec:robust-localizer}), $N^{BB}$ positions are sampled around the predicted position and IoUNet is applied to produce refined bounding boxes with predicted IoU scores. 
$N^{\mathrm{TOP}}$ bounding boxes with the highest predicted score are averaged to produce the final bounding box.

\subsection{Long-term tracker architecture}  \label{subsec:longterm}

A long-term tracker is required to address situation in which the target disappears for longer duration and re-appears later on. Target loss prediction and re-detection play a crucial role in these scenarios. We build on a single-model long-term tracking architecture~\cite{lukezicFCLT}.
In our case, the short-term tracker is composed of robust localizer i.e., a deep non-stationary DCF (Section~\ref{subsec:robust-localizer}) and an accurate bounding box refinement module (Section~\ref{subsec:fine-localizator}), and is used for continuous, short-term, target localization. Periods of unreliable target localization are detected by a depth-aware target presence classifier (Section~\ref{subsec:target-presence-classifier}). Once the target is deemed lost, the target search range progressively increases over the consecutive frames. Target is re-detected by applying the depth-modulated DCF from~\ref{subsec:robust-localizer} within the enlarged search region. 
Once the target is re-detected, the search range reduces back to that of short-term tracking.

Since the same model is used for short-term tracking and detection, care has to be taken to prevent model contamination and irrecoverable drift caused by updating from the background. We thus apply target presence indicators (Section~\ref{subsec:target-presence-classifier}) to switch between target presence/absence states and identify periods during which it is safe to update the target model.

\subsection{Depth-aware target presence indicators}  \label{subsec:target-presence-classifier}

The similarity between the model and the detected target is quantified by the maximum of the depth-modulated DCF correlation response, i.e., $\rho_\mathrm{DCF}$. Low value indicates a low target presence likelihood. Thus the correlation-based target presence indicator is defined as $\beta_\mathrm{DCF}(\tau) = \{ 1: \rho_\mathrm{DCF} > \tau ; 0:~\textrm{otherwise} \}$.

Temporal depth consistency is used as another indicator. The target is represented by the set of dept histograms $\mathcal{G}_i \in G, i=1,...,N_G$, extracted from the depth images from predicted bounding box region in the previous time-steps. A histogram extracted in the current time-step $\mathcal{H}$ is compared to these histograms by Bhattacharyya similarity
\begin{equation}  \label{eq:bha_depthhist}
    \rho_\mathrm{dep}^{i} = \sum_{j}^{n_B}\sqrt{\mathcal{H}_j\mathcal{G}_{j}^{i}},
\end{equation}
where $n_B$ is number of the histogram bins. Low values
indicate target occlusion or disappearance.
The depth consistency indicator is therefore defined as \mbox{$\beta_\mathrm{dep}(\tau) = \{ 1: \rho_\mathrm{dep}^{i} > \tau~\forall~ i; 0:~\textrm{otherwise} \}$}. The set of depth histograms is refreshed each time a target model is updated by first-in-first-out mechanism.

The correlation and depth consistency indicators are applied to construct conditions to trigger (i) target lost state, (ii) target re-detected state, and (iii) to decide whether it is safe to update the target model without background contamination. The triggers are summarized in Table~\ref{tab:actions_conditions}.

\begin{table}
\caption{Summary of the tracking state triggers in
Section~\ref{subsec:target-presence-classifier}. 
}
\label{tab:actions_conditions}
\centering
\begin{tabularx}{\linewidth}{lL} 
\textbf{State} & \textbf{Conditions} \\ 
\hline
Target lost & 
$\textit{cond}_1:  1-\beta_\mathrm{DCF}(\tau_l)$ \newline
$\textit{cond}_2:~1-\beta_\mathrm{DCF}(\tau)$~\&~$1-\beta_\mathrm{dep}(\tau_D)$ 
\\
\hline
Target re-detected & 
$\textit{cond}_1:\beta_\mathrm{DCF}(\tau_h)$
$\textit{cond}_2:\beta_\mathrm{DCF}(\tau)~\&~\beta_\mathrm{dep}(\tau_D)$
\\ 
\hline
Update model& 
$\textit{cond}_1$: $\beta_\mathrm{DCF}(\tau_u)$~\&~ $\beta_\mathrm{dep}(\tau_D)$
\\
\hline
\end{tabularx}
\end{table}

\section{Experiments}
\label{sec:results}

\subsection{Implementation Details}

\begin{table*}[!t]
\begin{center}
\caption{Experiments on the Princeton Tracking Benchmark using the PTB protocol. Results and ranks are retrieved from the online server, where our submission uses the nickname ``zoom2track'' on the website. For overall success rate and each tagged attribute, we annotate the top three methods. Numbers in the parenthesis are the ranks.
% The red text ``DA'' refers to the proosed depth-aware DCF, while ``LT'' means we long-termlize the base tracker.
}
\label{table:resultsPTB}
\scalebox{.80}
{
\begin{tabular}{lr llll llll llll}
%\toprule & \multicolumn{11}{c}{{\bf Attributes}}\\
\toprule
\specialcell{\bf Method} & \specialcell{\em } & \specialcell{\em Avg.Success} & \specialcell{\em Human} & \specialcell{\em Animal} & \specialcell{\em Rigid} & \specialcell{\em Large} & \specialcell{\em Small} & \specialcell{\em Slow} & \specialcell{\em Fast} & \specialcell{\em Occ.} & \specialcell{\em No-Occ.} & \specialcell{\em Passive} & \specialcell{\em Active}\\
\midrule
% \textit{\textit{DiMP+\textcolor{red}{DA}}} 
DAL & -- & {0.807}(1) & {0.78}(2) & {0.86}(1) & {0.81}(2) & {0.76} & {0.84}(1) & {0.83}(2) & {0.80}(1) & {0.72}(2) & {0.93}(1) & 0.78 & {0.82}(1) \\
\textit{OTR} \cite{kart2019object} & -- & {0.769}(2) & {0.77}(3) & 0.68  & 0.81(2) & 0.76 & {0.77}(3) & 0.81 & {0.75}(2) & 0.71 & 0.85 & {0.85}(1) & 0.74\\
\textit{DiMP}~\cite{dimp} & -- & 0.765(3) & 0.67 & {0.86}(1) & 0.79 & 0.67 & {0.81}(2) & {0.82}(3) & 0.73 & 0.63 & {0.93}(1) & 0.74 & {0.76}(2)  \\
\textit{ca3dms+toh}~\cite{ca3dms} & -- & 0.737  & 0.66 & 0.74 & {0.82}(1) & 0.73 & 0.74& 0.80& 0.71& 0.63& {0.88}(3) & {0.83}(2) & 0.70\\
\textit{CSR-rgbd++}~\cite{Kart_ECCVW} & -- & 0.740 & {0.77} & 0.65 & 0.76& 0.75& 0.73& 0.80 & 0.72 & 0.70 & 0.79 & 0.79 & 0.72\\
% \textit{ATOM}~\cite{atom} & -- & 0.745 & 0.61 & {0.81} & {0.83} & 0.62 & {0.81} & 0.81& 0.70 & 0.59 & {0.92} & 0.79 & 0.71 \\
\textit{3D-T}~\cite{Bibi3D} & -- & 0.750 & {0.81}(1) & 0.64 & 0.73 & {0.80}(1) & 0.71 & 0.75 & {0.75}(2) & {0.73}(1) & 0.78 & 0.79 & 0.73\\
\textit{PT}~\cite{princetonrgbd} & -- & 0.733 & 0.74 & 0.63 & 0.78 & 0.78(3) & 0.70 & 0.76 & 0.72 & 0.72(2) & 0.75 & 0.82(3) & 0.70\\
\textit{OAPF}~\cite{MESHGI_OAPF} & -- & 0.731 & 0.64 & {0.85}(3) & 0.77 & 0.73 & 0.73 & {0.85}(1) & 0.68 & 0.64 & 0.85 & 0.78 & 0.71\\
\textit{DLST}~\cite{DLST} & -- & 0.740 & {0.77} & 0.69 & 0.73 & {0.80}(1) & 0.70 & 0.73 & 0.74 & 0.66 & 0.85 & 0.72 & {0.75}(3)\\
\textit{DM-DCF}~\cite{DMDCF} & -- & 0.726 & 0.76 & 0.58 & 0.77 & 0.72 & 0.73 & 0.75 & 0.72 & 0.69 & 0.78 & 0.82 & 0.69\\
\textit{DS-KCF-Shape}~\cite{Hannuna2016} & -- & 0.719 & 0.71 & 0.71 & 0.74 & 0.74 & 0.70 & 0.76 & 0.70 & 0.65 & 0.81 & 0.77 & 0.70\\
\textit{DS-KCF}~\cite{dskcf_bmvc} & -- & 0.693 & 0.67 & 0.61 & 0.76 & 0.69 & 0.70 & 0.75 & 0.67 & 0.63 & 0.78 & 0.79 & 0.66\\
\textit{DS-KCF-CPP}~\cite{Hannuna2016} & -- & 0.681 & 0.65 & 0.64 & 0.74 & 0.66 & 0.69 & 0.76 & 0.65 & 0.60 & 0.79 & 0.80 & 0.64 \\
\textit{hiob-lc2}~\cite{hiob} & -- & 0.662 & 0.53 & 0.72 & 0.78 & 0.61 & 0.70 & 0.72 & 0.64 & 0.53 & 0.85 & 0.77 & 0.62\\
\textit{STC}~\cite{stc} & -- & 0.698 & 0.65 & 0.67 & 0.74 & 0.68 & 0.69 & 0.72 & 0.68 & 0.61 & 0.80 & 0.78 & 0.66\\
\bottomrule
\end{tabular}
}
\end{center}
\end{table*}

The backbones for deep DCF and the IoUNet are pre-trained for localization task on RGB sequences and the filter update parameters are kept as in~\cite{dimp}.
The depth modulation hyperparameter is set to $\alpha = 0.1$.  The  binds in depth histograms are constrained to 8 meters at resolution of 0.1m per bin. The search region enlargement rate factor during re-detection is set to $r=1.05$.% and limited to the double of the initial search region size.
The target presence indicator thresholds (Table~\ref{tab:actions_conditions}) are set to $\beta_l = 0.2$, $\beta = 0.25$, $\beta_h = 0.3$ and $\beta_D = 0.8$.
The number of depth histograms in the depth temporal consistency model is set to $N_G = 3$.
The preliminary study showed that the method is not sensitive to these parameters and we use the same values in all experiments.

\subsection{State-of-the-art Comparison}

The proposed depth-aware long-term (DAL) tracker is evaluated on three major RGB-D benchmarks: 
Princeton tracking benchmark~\cite{princetonrgbd} (PTB), STC tracking benchmark~\cite{stc} and Color-and-depth tracking benchmark~\cite{cdtb} (CDTB).
In the following we discuss the tracking performance on these benchmarks.

\subsubsection{Princeton Tracking Benchmark}

Princeton Tracking Benchmark~\cite{princetonrgbd} (PTB) is the most popular benchmark in RGBD tracking.
The dataset consists of 95 video sequences without publicly available ground-truth annotations to prevent over-fitting.
The sequences are annotated with 11 visual attributes  for a thorough analysis of tracking performance (Table~\ref{table:resultsPTB}). 
A per-frame tracking performance is measured by a modified overlap measure that sets the overlap to 1 on frames where the target is correctly predicted to be missing.
The overall tracking performance is measured by the \textit{success rate}, i.e., the percentage of frames where overlap between ground truth and the predicted bounding box exceeds 0.5.

All state-of-the-art RGBD trackers from PTB are included in the analysis: OTR~\cite{kart2019object}, ca3dms+toh~\cite{ca3dms}, CSR-rgbd++~\cite{Kart_ECCVW}, 3D-T~\cite{Bibi3D}, PT~\cite{princetonrgbd}, OAPF~\cite{MESHGI_OAPF}, DM-DCF~\cite{DMDCF}, DS-KCF-Shape~\cite{Hannuna2016}, DS-KCF~\cite{dskcf_bmvc}, DS-KCF-CPP~\cite{Hannuna2016}, hiob\_lc2~\cite{hiob}, STC~\cite{stc} and DLST~\cite{DLST}. 
We additionally include the state-of-the-art short-term RGB tracker DiMP~\cite{dimp}, which ranks top on the most short-term tracking benchmarks.

\begin{figure}[t]
  \begin{center}
    \includegraphics[width=\linewidth]{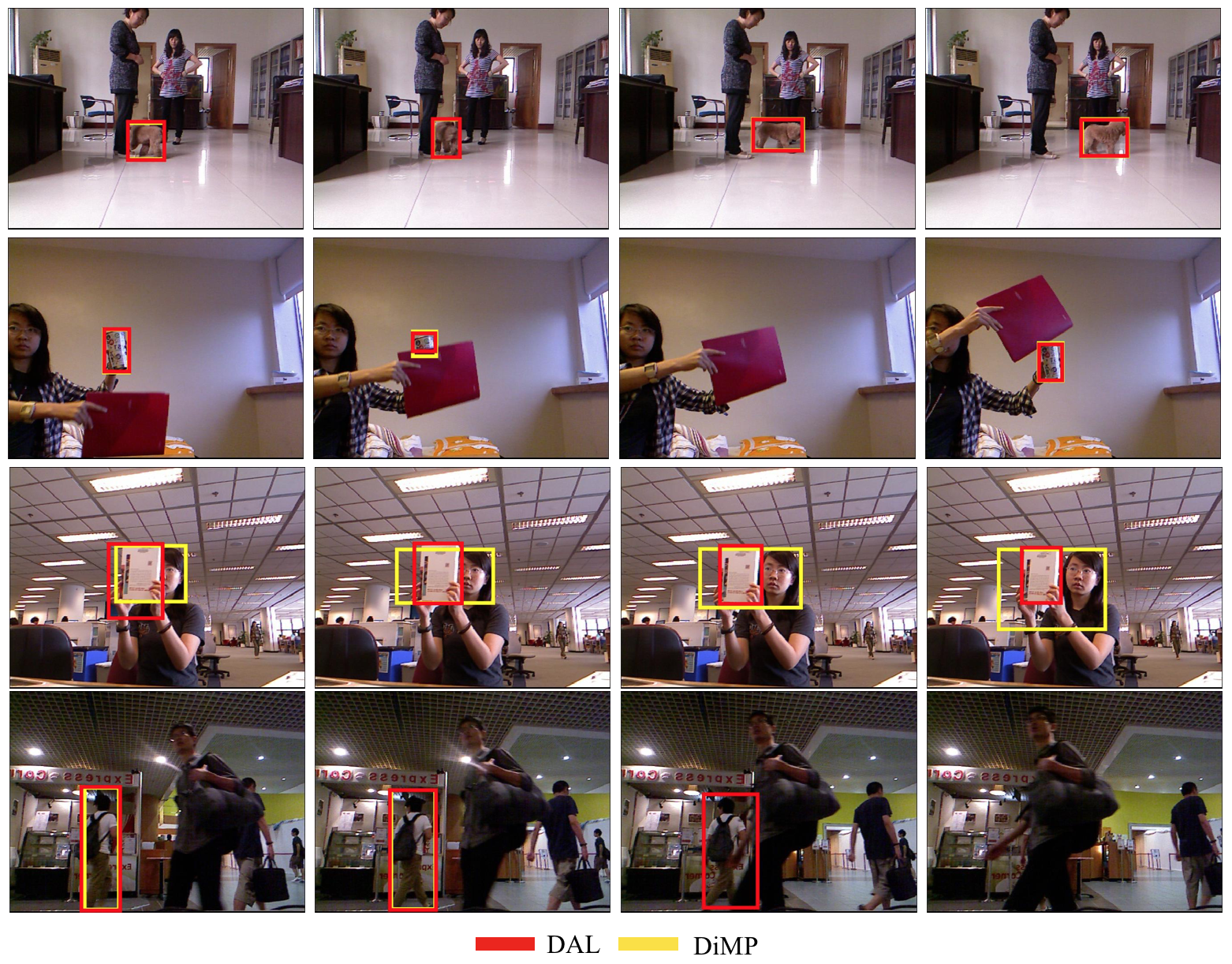}
    \caption{
    Qualitative comparison of DAL and DiMP. Both trackers localizes the target and give precise bounding boxes (the first two rows). With depth-modulated DCF, our tracker shows better discriminative ability when strong distractor appears (human face in the third row).
    Compared to DiMP, with conservative long-term tracking design, our tracker reports target disappearance more accurately.
    }
      \label{fig:ptbcompare}
  \end{center} 
\end{figure}

DAL achieves the average success rate higher than 0.8, outperforming all RGBD trackers and outperform the sota RGBD tracker OTR and the sota RGB tracker DiMP by 5\%. On most attributes except ``Passive motion'', DAL ranks the first or the second, showing its robustness under various tracking conditions.
Compared to OTR, the success rates on ``Animal, Small Object, No-Occlusion, Active Motion'' are significantly better, verifying the improved utilization in depth for tracking. The per-attribute results also show that DAL deals better with occlusion than DiMP, which may be attributed to the nonstationary DCF modulated by the depth map.
See Figure~\ref{fig:ptbcompare} for qualitative comparison on PTB.

\begin{table*}
\scriptsize
\begin{center}
  \caption{The normalized area under the curve (AUC) scores computed from one-pass evaluation on the STC Benchmark~\cite{stc}. Top three results for each attributes are annotated. 
  }
\label{table:resultsSTC}\scalebox{1.1}
{
{\begin{tabular}{llllllllllll}
\toprule \multicolumn{12}{c}{\bf Attributes}\\
{\bf Method} & {\em AUC} & {\em IV} & {\em DV} & {\em SV} & {\em CDV} & {\em DDV} & {\em SDC} & {\em SCC} & {\em BCC} & {\em BSC} & {\em PO}\\
\midrule
DAL & {0.64}(1) & {0.51}(1) & {0.63}(1) & {0.50}(1) & {0.60}(1) & {0.62}(1) & {0.64}(1) & {0.63}(2) & {0.57}(1) & {0.58}(1) & {0.58}(1)\\
\textit{DiMP}~\cite{dimp} & {0.61}(2) & {0.50}(2) & {0.62}(2) & {0.48}(2) & {0.57}(2) & {0.58}(2) & {0.61}(2) & {0.65}(1) & {0.52}(2) & {0.55}(2) & {0.58}(1)\\
\textit{OTR}~\cite{kart2019object} & {0.49}(3) & {0.39}(3) & {0.48}(3) & {0.31}(3) & 0.19 & {0.45}(3) & {0.44}(3) & 0.46 & {0.42}(3) & {0.42}(3) & {0.50}(3)\\
\textit{CSR-rgbd++}~\cite{Kart_ECCVW} & 0.45 & 0.35 & 0.43 & 0.30 & 0.14 & 0.39 & 0.40 & 0.43 & 0.38 & 0.40 & 0.46\\
\textit{ca3dms+toh}~\cite{ca3dms} & 0.43 & 0.25 & 0.39 & 0.29 & 0.17 & 0.33 & 0.41 & {0.48}(3) & 0.35 & 0.39 & 0.44\\
\textit{STC}~\cite{stc} & 0.40 & 0.28 & 0.36 & 0.24 & {0.24}(3) & 0.36 & 0.38 & 0.45 & 0.32 & 0.34 & 0.37\\
\textit{DS-KCF-Shape}~\cite{Hannuna2016} & 0.39 & 0.29 & 0.38 & 0.21 & 0.04 & 0.25 & 0.38 & 0.47 & 0.27 & 0.31 & 0.37\\
\textit{PT}~\cite{princetonrgbd} & 0.35 & 0.20 & 0.32 & 0.13 & 0.02 & 0.17 & 0.32 & 0.39 & 0.27 & 0.27 & 0.30\\
\textit{DS-KCF}~\cite{dskcf_bmvc} & 0.34 & 0.26 & 0.34 & 0.16 & 0.07 & 0.20 & 0.38 & 0.39 & 0.23 & 0.25 & 0.29\\
\textit{OAPF}~\cite{MESHGI_OAPF} & 0.26 & 0.15 & 0.21 & 0.15 & 0.15 & 0.18 & 0.24 & 0.29 & 0.18 & 0.23 & 0.28\\
\bottomrule
\end{tabular}}}
\end{center}
\end{table*}

\subsubsection{STC Tracking Benchmark}

The STC benchmark~\cite{stc} is complementary to the PTB in terms of visual attributes and 
contains 36 sequences annotated per-frame with 10 attributes: \textit{ Illumination variation} (IV), \textit{Depth variation} (DV), \textit{Scale variation} (SV), \textit{Color distribution variation} (CDV), \textit{Depth distribution variation} (DDV), \textit{Surrounding depth clutter} (SDC), \textit{Surrounding color clutter} (SCC), \textit{Background color camouflages} (BCC), \textit{Background shape camouflages} (BSC), \textit{Partial occlusion} (PO).

Since the targets in STC dataset are always visible, the standard short-term tracking evaluation methodology is used~\cite{wu2015object}. 
Tracking performance is evaluated by the success and precision plots.
The success plot shows percentage of frames where overlap of the predicted bounding box is larger than a threshold, for a set of overlap thresholds.
Trackers are ranked according to the area under the success rate curve.
The precision plot shows percentage of frames where distance between the predicted bounding box center and the ground-truth bounding box center is smaller than the threshold, for a set of center error thresholds. 
Trackers are ranked according to the performance at the threshold of 20 pixels.

The proposed tracker is compared to the following state-of-the-art RGBD trackers:
OTR~\cite{kart2019object}, CSR-rgbd++~\cite{Kart_ECCVW}, ca3dms+toh~\cite{ca3dms}, STC~\cite{stc}, DS-KCF-Shape~\cite{Hannuna2016}, PT~\cite{princetonrgbd}, DS-KCF~\cite{dskcf_bmvc} and OAPF~\cite{MESHGI_OAPF}.
The most recent sota RGB short-term tracker (DiMP~\cite{dimp}) is included as well.

\begin{figure}[ht]
  \begin{center}
    \includegraphics[width=1.05\linewidth]{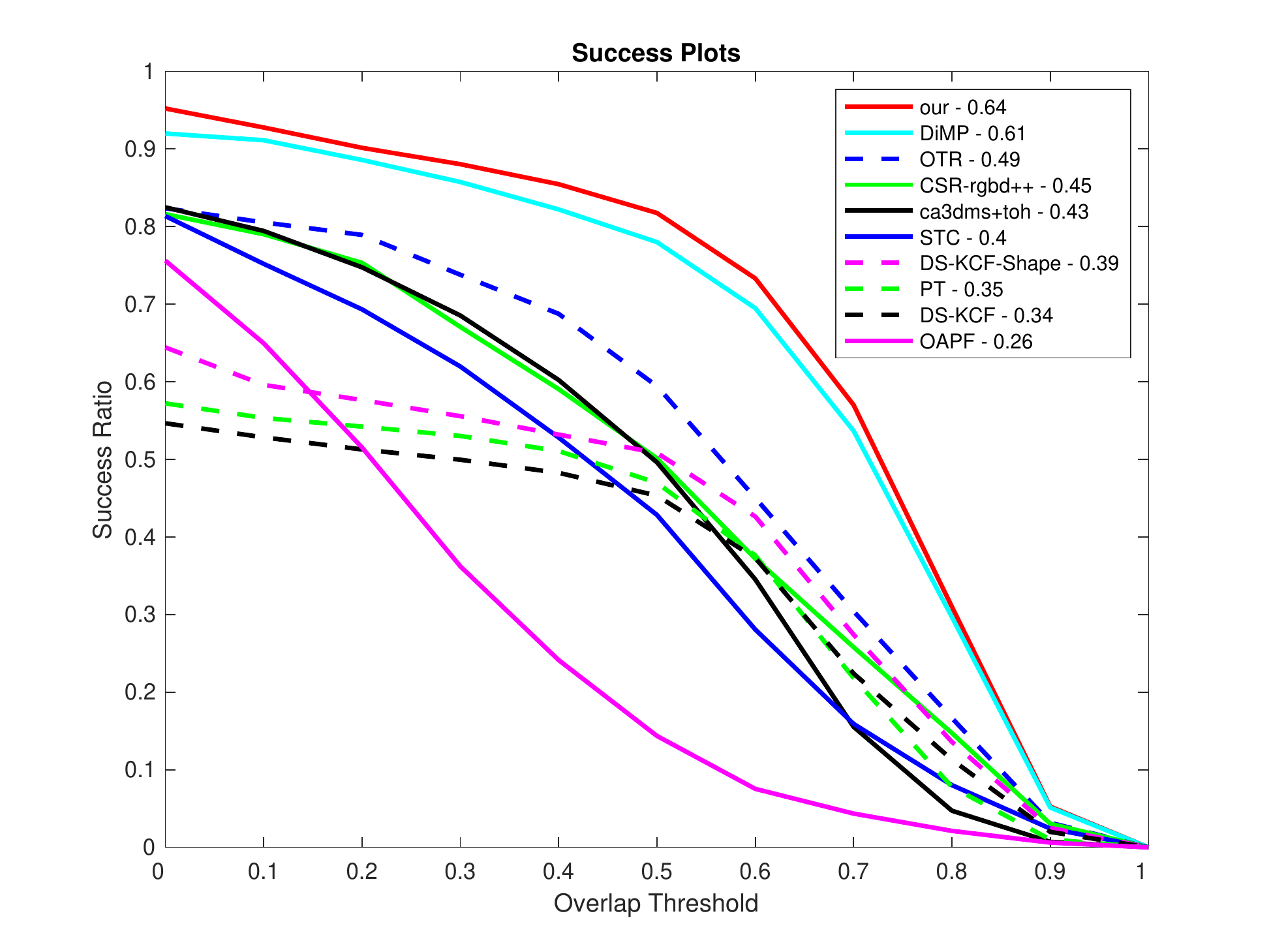}
    %\caption{Success and precision plots on STC benchmark~\cite{STC}.}
    %\label{fig:STC_AUC_Precision}
    \includegraphics[width=1.05\linewidth]{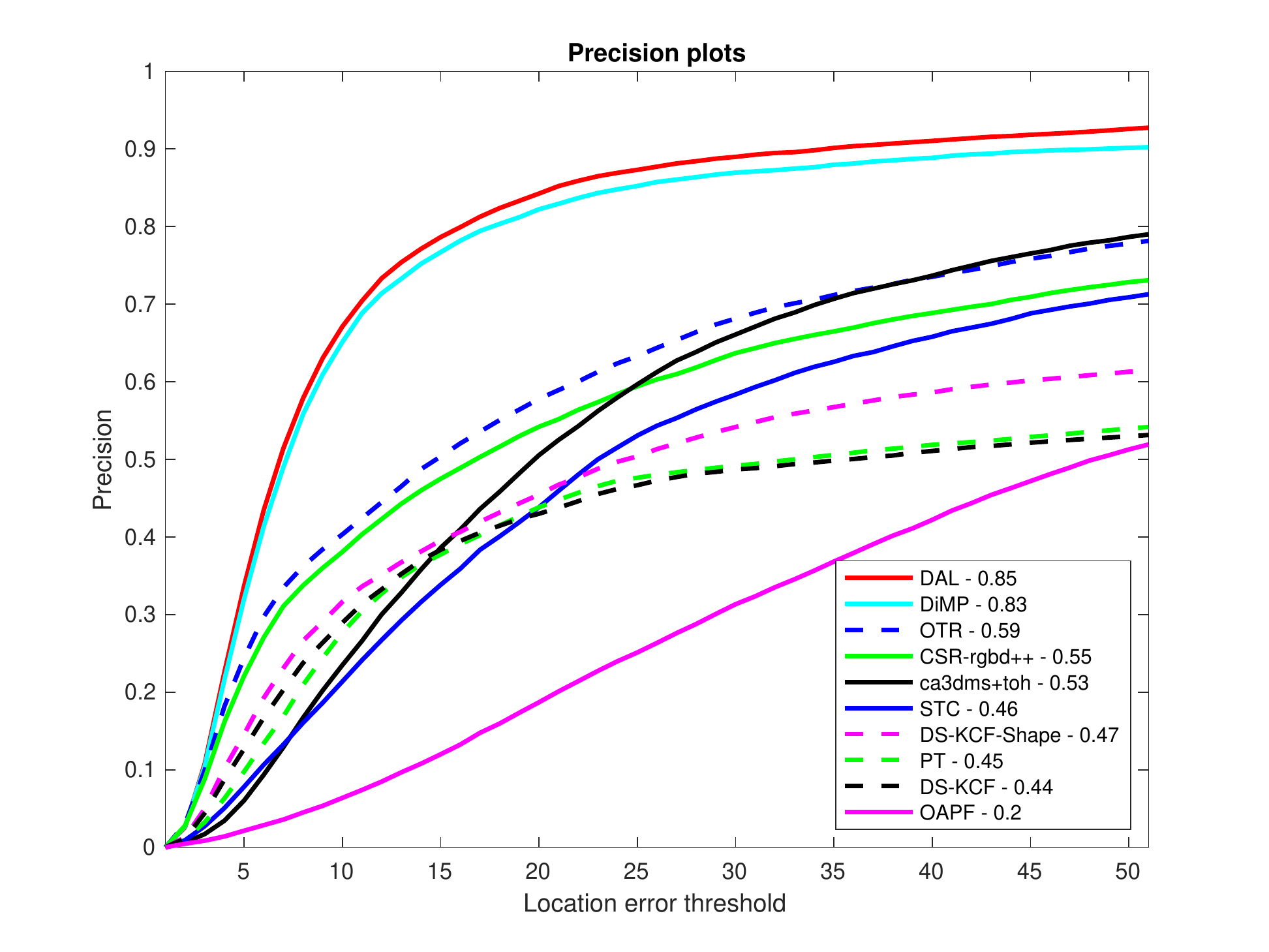}
    \caption{Success and precision plots on STC benchmark~\cite{stc}.}
      \label{fig:STC_AUC_Precision}
  \end{center}
\end{figure}

Results are reported in Figure~\ref{fig:STC_AUC_Precision}. DAL outperforms top-performing RGBD trackers by a large margin. The top RGBD tracker OTR is outperformed significally by 30.6\%, while DiMP is outperformed by 4.9\%. 
The improved performance is consistent across all the attributes, except SCC (Surrounding Color Clutter).

\begin{figure}
\centering
\includegraphics[width=1.1\linewidth]{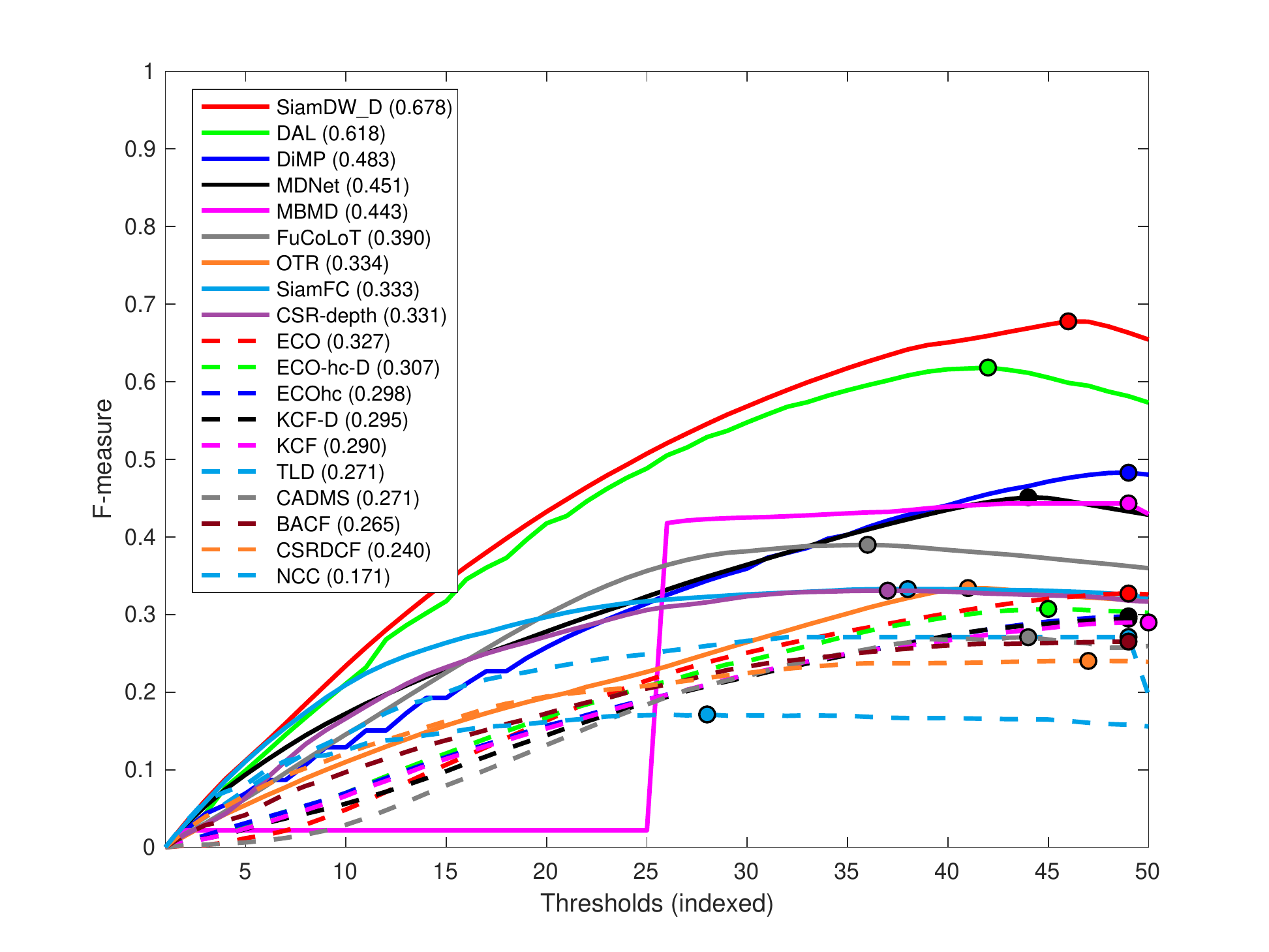}
\includegraphics[width=1.1\linewidth]{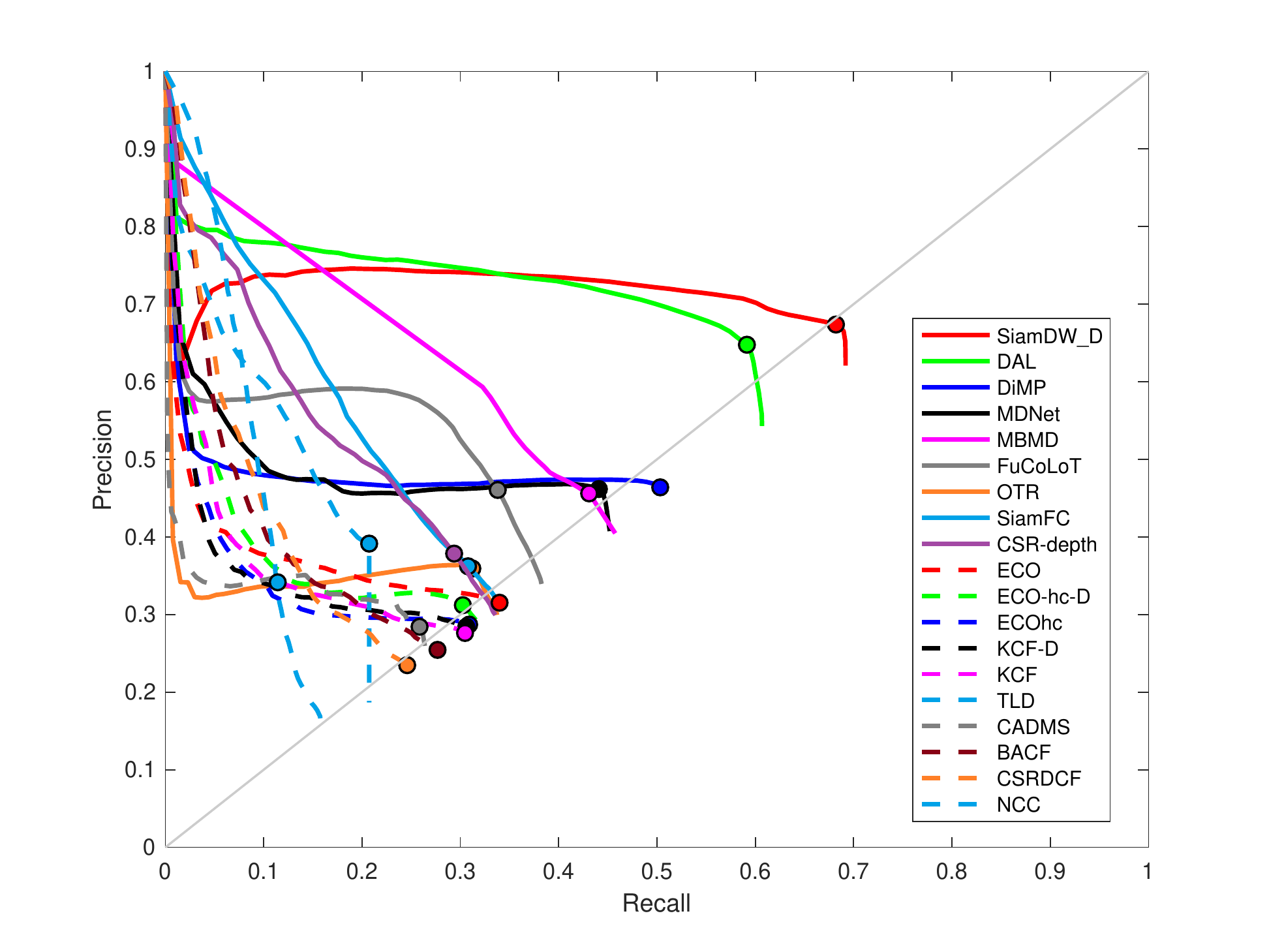}
\caption{The overall tracking performance is presented as tracking F-measure (top) and tracking Precision-Recall (bottom). Trackers are ranked by their optimal
tracking % can be deleted to optimize length
performance (maximum F-measure).}
\label{fig:overall-performance}
%\vspace{-11mm}
\end{figure}

\subsection{Color and depth tracking benchmark}

The CDTB dataset~\cite{cdtb} is the most recent and the most challenging RGBD dataset. The sequences are captured in the long-term tracking scenario, which means that the target is often fully occluded or that it disappears from the field of view.
The most important aspects in long-term tracking are therefore ability to predict target absence and target re-detection.
Tracking performance is measured as tracking recall ($Re$, average overlap on frames where the target is visible) and tracking precision ($Pr$, average overlap on frames where tracker makes a prediction). 
Trackers are ranked according to the tracking F-measure, which is combination of $Pr$ and $Re$.

The proposed tracker is compared to all top trackers from the CDTB benchmark: 
(i) sota short-term RGB trackers (KCF~\cite{Henriques_KCF}, NCC~\cite{Kristan_VOT2017}, BACF~\cite{bacf}, CSRDCF~\cite{lukezic2017discriminative}, SiamFC~\cite{SiameseFC}, ECOhc~\cite{ECO}, ECO~\cite{ECO} and MDNet~\cite{mdnet}), 
(ii) sota long-term RGB trackers (TLD~\cite{tld}, FuCoLoT~\cite{lukezicFCLT} and MBMD~\cite{mbmd}) and 
(iii) sota RGBD trackers (OTR~\cite{kart2019object}, Ca3dMS~\cite{ca3dms}, ECOhc-D~\cite{Kart_ECCVW} and CSRDCF-D~\cite{Kart_ECCVW}). 
We also include the most recent short-term RGB tracker DiMP~\cite{dimp} which is the top-performer on the most of the short-term datasets and the winner of the recent VOT2019 RGBD challenge~\cite{Kristan_2019_ICCV} (SiamDW-D~\cite{Kristan_2019_ICCV}).

Tracking results are presented in Figure~\ref{fig:overall-performance}.
The proposed tracker outperforms the top-performer in CDTB~\cite{cdtb}, MDNet, by a large margin of 37\% mostly due to the powerful re-detection module and the non-stationary DCF.
The OTR, which is the sota RGBD tracker, is outperformed by 85\% mostly due to the better target representation including deep features and the deep non-stationary DCF.
The proposed tracker outperforms sota RBG short-term DiMP \wrt the all three measures: tracking F-measure by 28\%, precision by 40\% and recall by 18\%, which demonstrates the impact of the re-detection component and the non-stationary DCF.
The top-performing tracker in VOT2019 RGBD challenge, SiamDW-D slightly outperforms DAL.
%the proposed tracker by 9.7\%.
SiamDW-D is a complex combination of multiple short-term tracking methods and general object detectors from the off-the-shelf toolbox~\cite{fpn}. This complicated architecture does prevents significant incorporation of depth in the tracker. In fact, depth is used only for target loss identification. Due to computational complexity, SiamDW-D performs at very low frame rate (2 fps as we test) and has large memory footprint due to several network branches. On the other hand, DAL has a very streamlined trainable architecture and runs 10$\times$ faster thanks to efficient use of depth, while attaining comparable tracking accuracy.

\begin{figure}
\centering
\includegraphics[width=1.1\linewidth]{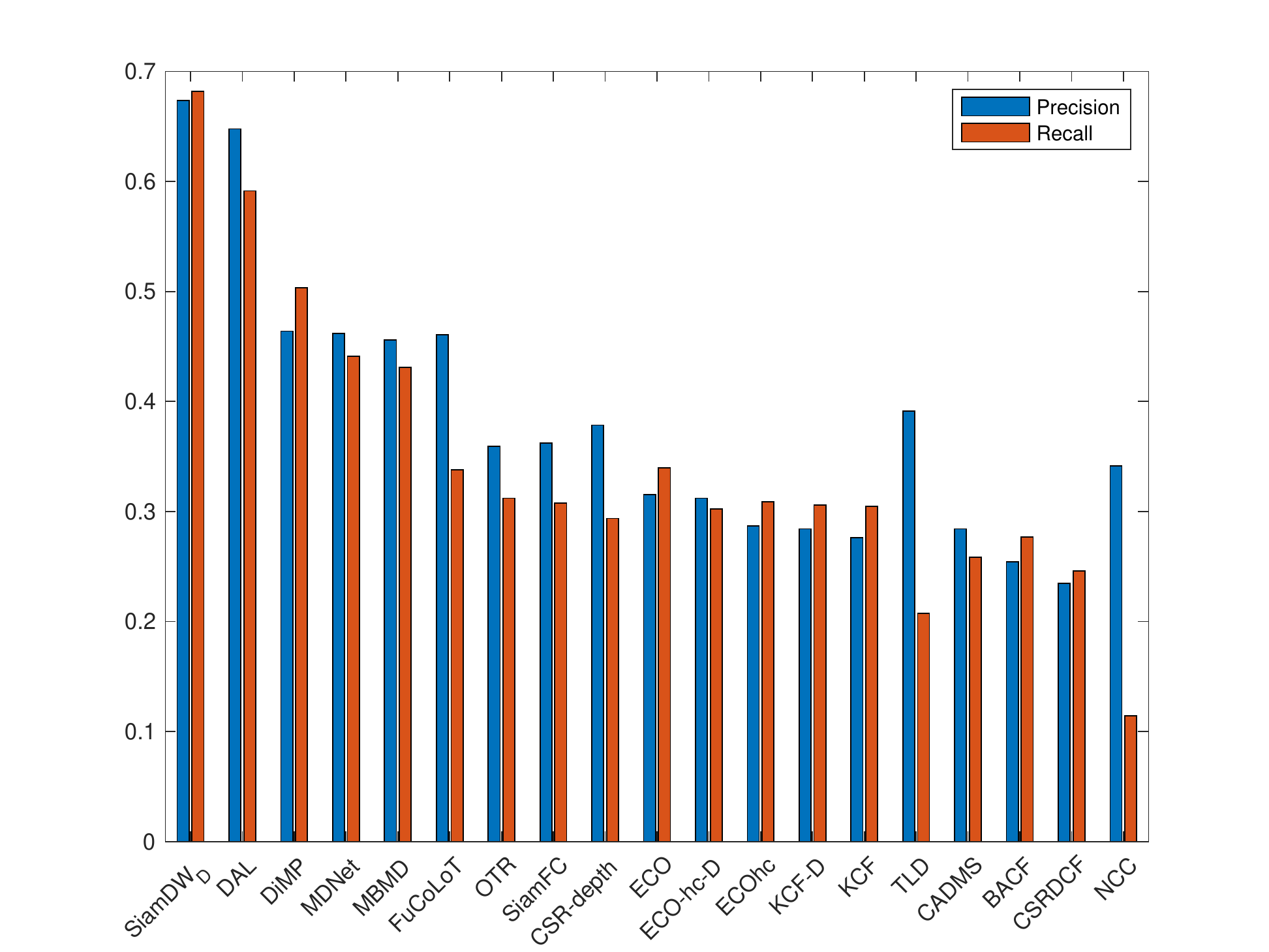}
\caption{Tracking precision and recall calculated at the optimal point (maximum F-measure).}
\label{fig:overall-precision-recall}
\end{figure}

\subsection{Ablation Study}  \label{sec:ablation}

\begin{figure}
\centering
\includegraphics[width=1.1\linewidth]{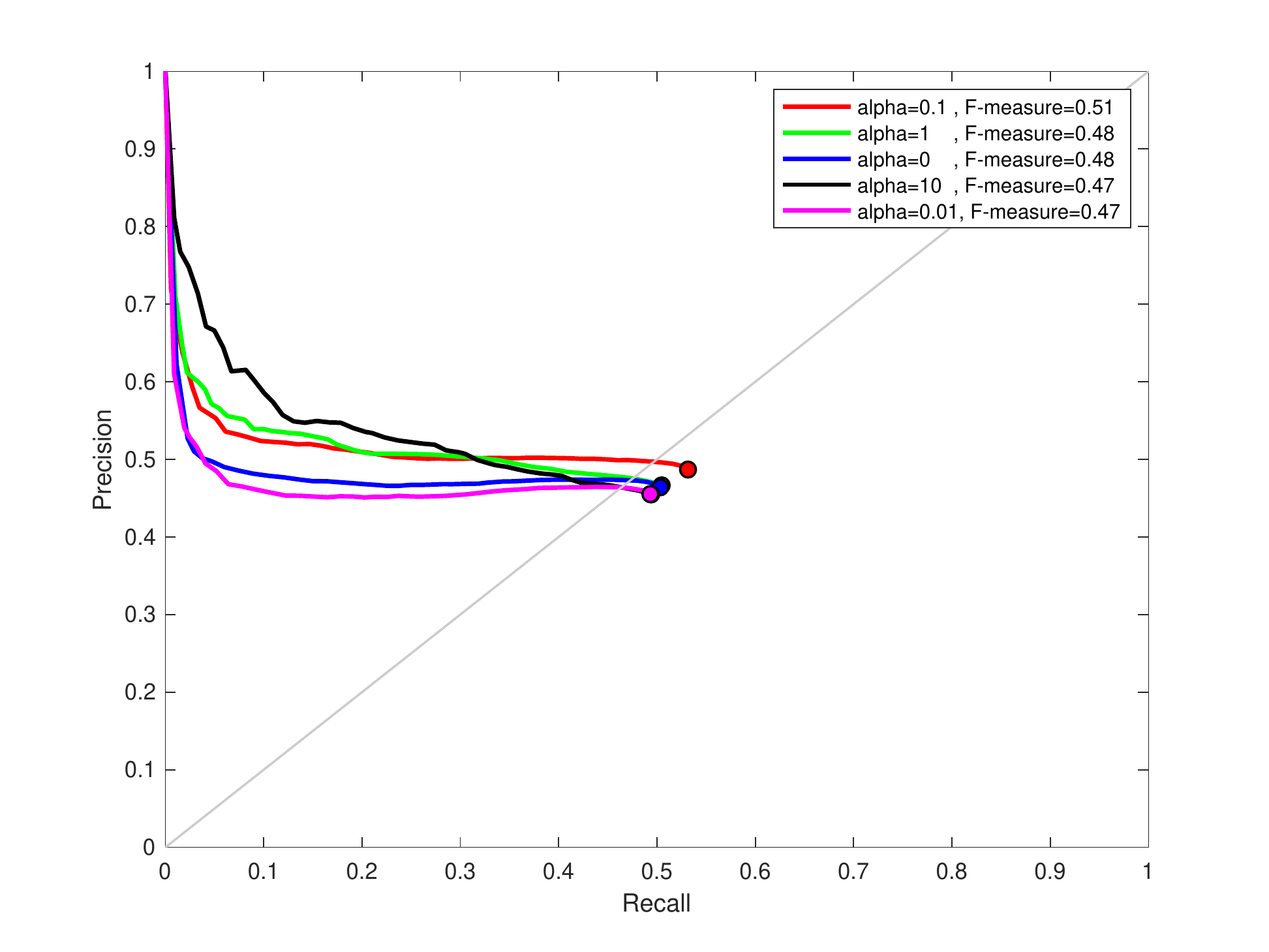}
\caption{Precision-Recall curves and F-measure as function of varying $\alpha.$ for depth-modulated DCF. Evalauted on CDTB dataset.}
\label{fig:votd_ablation}
\end{figure}

An ablation study was conducted on the most challenging RGBD dataset~\cite{cdtb} to demonstrate the contribution of each component of DAL.

The following variants of DAL are evaluated: 
(i) DAL, the proposed method uses the non-stationary DCF and target presence classifier using DCF confidence and depth, described in Section~\ref{subsec:target-presence-classifier}, to activate the re-detection, is denoted as +$\alpha$+LT$^{\beta,\beta_{\mathrm{D}}}$.
(ii) DAL$^{- LT(\beta_D)}$, $\beta_D$ is not considered in computing target presence.
(iii) DAL$^{-\alpha,- LT(\beta_D)}$, $\beta_D$ is not considered in computing target presence and only base DCF is used.
(iv) DAL$^{- LT}$, long-term design is turned off, depth-modulated DCF is used.
(v) DAL$^{-\alpha,- LT}$, long-term design is turned off and only base DCF is used, which equals to the base short-term tracker.

The results are shown in the Table~\ref{tab:ablation_lt}. The short-term version of DAL using a stationary DCF, DAL$^{-\alpha,- LT}$,  achieves 0.48 F-measure.
Adding non-stationary formulation of DCF in DAL$^{- LT}$ improves the results for 6.3\%. The result shows that correlation-based trackers can benefit from the non-stationary DCF formulation.
The baseline tracker with a stationary DCF extended to the long-term scenario (DAL$^{-\alpha,- LT(\beta_D)}$) improves the results for 14.6\%, which shows the importance of re-detection in long-term tracking scenario.
Combining both, non-stationary DCF formulation and target re-detection (DAL$^{- LT(\beta_D)}$) improves the results for 20.8\%.
Finally, the performance boost of 29.1\% is achieved by the non-stationary DCF formulation and target re-detection activated by the multiple conditions using DCF confidence and depth (DAL).
This result shows that depth can significantly improve tracking performance. 

We additionally performed a sensitivity study of the parameter $\alpha$ from (\ref{eq:depthdisparity}), which controls the modulation strength in the DCF depth modulation map.
The baseline tracker (i.e., without depth modulation, $\alpha = 0$) was extended by the non-stationary DCF formulation (without target re-detection) and the following values of $\alpha$ were tested: 0.01, 0.1, 1 and 10.
The results in Figure~\ref{fig:votd_ablation} show that the highest performance is obtained at $\alpha = 0.1$. Lower $\alpha$  push tracking performance to the baseline tracker.
Increasing $\alpha$ amplifies the depth modulation too much, causing a slight performance drop.

\begin{figure}
\centering
\includegraphics[width=\linewidth]{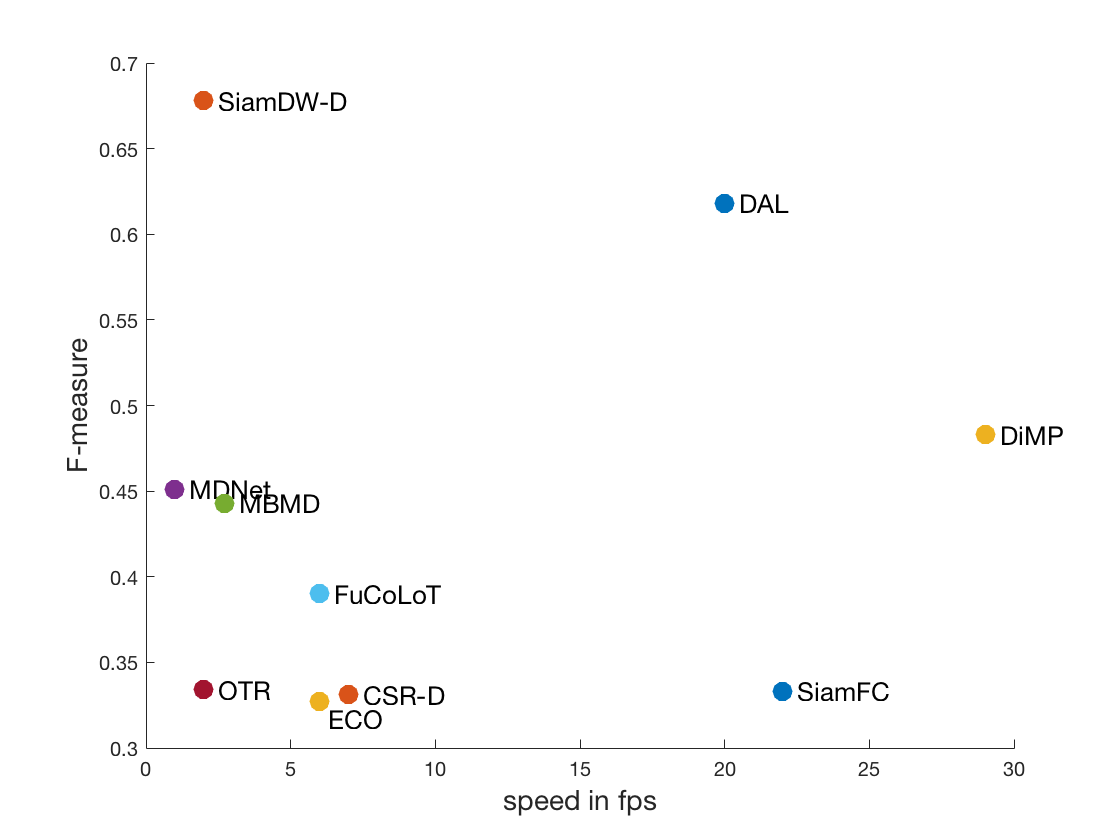}
\caption{Tracker practicality evaluation with respect to  F-measuare and Speed (in frames-per-second).}
\label{fig:speed}
\end{figure}

{\bf Tracking speed}. 
We measure the speed of ten top-performing trackers on the CDTB dataset to evaluate the performance in the context of practical applications that require accurate tracking at high speeds, i.e., robotics and real-time systems. Results are shown in Figure~\ref{fig:speed}.
DAL runs close to the real-time, at 20 frames per second, while most of the other trackers (MDNet, MBMD, OTR, ECO, CSR-D) are much slower and  achieve significantly lower tracking accuracy.
SiamFC runs similarly fast to DAL, but it achieves 46.1\% lower tracking performance, DiMP is 45.0\% faster, but it achieves 21.8\% lower F-measure.
The top-performing SiamDW-D achieves 9.7\% higher F-measure, but it is 10-times slower. Thus DAL is the top-performing tracker in accuracy among the close-to-realtime tracking.

\begin{table}
\caption{DAL ablation study on the CDTB showing tracking F-measure.}
\label{tab:ablation_lt}
\scriptsize
\centering
\begin{tabular}{c c c c c} 
\hline
 DAL$^{-\alpha,- LT}$ & DAL$^{- LT}$ & DAL$^{-\alpha,- LT(\beta_D)}$ & DAL$^{- LT(\beta_D)}$ & DAL \\
\hline
0.48 & 0.51 & 0.55 & 0.58 & 0.62\\
\hline
\end{tabular}
\end{table}

\section{Conclusions}

We propose a novel deep DCF formulation for RGBD tracking. The formulation embeds depth information into the correlation filter optimization and provides a strong short-term RGBD tracker, improving the performance from 5\% to 6\% on all RGBD tracking benchmarks. 
We also propose a long-term tracking architecture where the
same deep DCF is used in target re-detection and depth
based tests effectively trigger between the short-term tracking,
re-detection and model update modes.
The long-term tracker consistently achieves
superior performance over the state-of-the-art RGB and
RGBD trackers (DiMP and OTR) on all three available RGBD
tracking benchmarks (PTB, STC and CDTB) and runs significantly
faster than the best RGBD competitor (2 fps vs. 20 fps).

{\small
\bibliographystyle{ieee_fullname}
\bibliography{bib}
}

\end{document}